\documentclass[conference]{IEEEtran}
\IEEEoverridecommandlockouts
\usepackage{cite}
\usepackage{amsmath,amssymb,amsfonts}
\usepackage{algorithmic}
\usepackage{graphicx}
\usepackage{textcomp}
\usepackage{xcolor}

\usepackage{amsmath} 
\usepackage{amssymb} 
\usepackage{graphicx}
 \usepackage[caption=false, font=normalsize, labelfont=sf, textfont=sf]{subfig}
\usepackage{multirow}
\usepackage{bm}

\newcommand{\RMDE}{RMDE}
\newcommand{\PVL}{PVL}
\newcommand{\CQP}{CQP}

\def\BibTeX{{\rm B\kern-.05em{\sc i\kern-.025em b}\kern-.08em
    T\kern-.1667em\lower.7ex\hbox{E}\kern-.125emX}}

\begin{document}

\title{DPA-I2P: Depth-Guided Projective Alignment for Image-to-Point-Cloud Registration in Autonomous Driving\\}

\author{
\IEEEauthorblockN{
Wenxin Zhang\IEEEauthorrefmark{1},
Hang Li\IEEEauthorrefmark{1},
Zhiwei Xu\IEEEauthorrefmark{2},
Qiankun Dong\IEEEauthorrefmark{1},
Gang Wang\IEEEauthorrefmark{1},
Tao Li\IEEEauthorrefmark{1}
}
\IEEEauthorblockA{
\IEEEauthorrefmark{1}Nankai University, Tianjin, China\\
wenxinzhang@mail.nankai.edu.cn,
lihangws2000@gmail.com,
\{qiankund, ganggang, litao\}@nankai.edu.cn
}
\IEEEauthorblockA{
\IEEEauthorrefmark{2}Haihe Lab of ITAI, Tianjin 300459, China\\
xuzhiwei2001@ict.ac.cn
}
}

\maketitle

\begin{abstract}
Image-to-Point Cloud Registration aims to estimate the camera pose of a given image within a 3D scene point cloud, which is a fundamental task in autonomous driving and large-scale outdoor localization. Recent implicit correspondence learning methods have improved registration performance by learning cross-modal alignment in an end-to-end framework, leading to more accurate camera pose estimation. However, due to the inherent modality discrepancy between images and sparse LiDAR point clouds, reliable cross-modal correspondence learning remains challenging. To address this issue, we propose Depth-Guided Projective Alignment for Image-to-Point-Cloud Registration (DPA-I2P). Unlike naive depth or feature concatenation, Ray-Conditioned Metric Depth Encoding (RMDE) and Projection-Consistent Vision Lifting (PVL) exploit depth and visual cues in a structured, geometry-aware manner. In addition, Cross-Modal Query Pruning (CQP) suppresses unreliable queries during early refinement to improve matching stability. Experiments on KITTI and nuScenes demonstrate the effectiveness of the proposed method. On KITTI, DPA-I2P reduces RTE and RRE by 45.0\% and 55.6\% over the strongest implicit baseline, respectively. On nuScenes, DPA-I2P also improves registration accuracy over the evaluated baselines, suggesting better transferability to different driving scenes.
\end{abstract}


\section{Introduction}
Image-to-Point Cloud (I2P) Registration aims to estimate the camera pose of an image within a 3D scene point cloud, which is a fundamental task for autonomous driving applications such as visual localization~\cite{pietrantoni2023segloc}, camera relocalization~\cite{moreau2023crossfire}, SLAM~\cite{lipson2024multisession}, motion planning~\cite{hu2023planning}, and 3D reconstruction~\cite{li2023neuralangelo}. Despite significant progress in recent years, accurate Image-to-Point Cloud Registration remains challenging due to the inherent modality gap between dense 2D image observations and sparse 3D point cloud representations.

Existing I2P methods mainly rely on explicit correspondence estimation~\cite{feng20192d3d,pham2020lcd,ren2022corri2p,zhou2023differentiable}, direct pose regression~\cite{li2021deepi2p,yao2024mafreei2p}, or implicit correspondence learning~\cite{li2025implicit,mu2025diff,cheng2025i2p}. Among them, implicit correspondence learning methods achieve stronger robustness by avoiding brittle hand-crafted matching and enabling end-to-end optimization. However, current methods still struggle to establish geometrically consistent cross-modal representations across heterogeneous modalities, especially in large-scale outdoor scenes with sparse LiDAR observations.

\begin{figure}[!t]

\centering

\includegraphics[width=\linewidth]{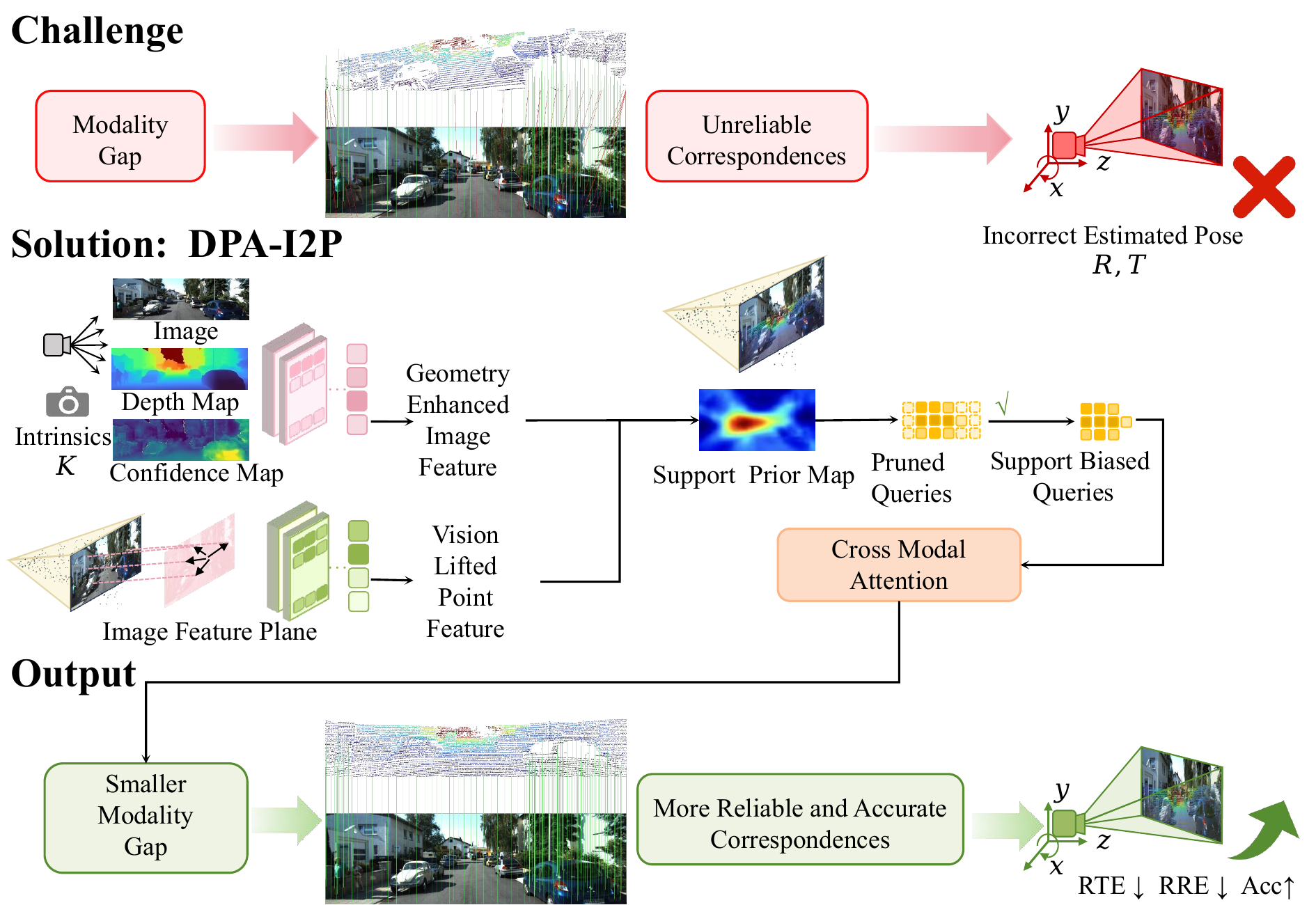}

\caption{Motivation of DPA-I2P. Existing implicit correspondence learning methods suffer from weak cross-modal feature consistency. DPA-I2P integrates depth and projection cues to construct a discriminative cross-modal feature space for more reliable correspondence exploration.}

\label{fig:motivation}

\end{figure}

The fundamental limitation lies in cross-modal feature inconsistency caused by the modality gap. Image features are often dominated by visual appearance without sufficient metric-scale geometric awareness, resulting in ambiguous matching in repetitive or low-texture regions. In contrast, point cloud features encode rich 3D geometric structures but lack reliable visual grounding, making cross-modal association unstable under coarse pose estimation. Consequently, the inconsistent feature space easily leads to correspondence ambiguity, which further introduces optimization instability during iterative pose refinement and degrades camera pose estimation accuracy. Therefore, the key challenge is to construct geometry-aware image representations and visually grounded point representations for reliable implicit correspondence learning (Fig.~\ref{fig:motivation}).

Accurate correspondence reasoning requires metric-scale depth awareness and direction cues to capture camera geometry. Recent monocular metric depth estimation methods, particularly UniDepthV2~\cite{piccinelli2025unidepthv2}, provide effective geometric priors for narrowing the modality gap between images and point clouds. However, metric depth alone is insufficient to explicitly represent visual geometry. Therefore, we consider metric depth, camera ray directions, and projection consistency to construct geometry-aware image representations and projection-consistent point features to improve cross-modal feature alignment and matching stability. In this way, we propose DPA-I2P, a depth-aware implicit correspondence learning framework for Image-to-Point Cloud Registration. Specifically, Ray-Conditioned Metric Depth Encoding (RMDE) incorporates depth cues into image representations to enhance image-side geometric awareness. Projection-Consistent Vision Lifting (PVL) enriches point representations with visual information. RMDE and PVL jointly construct a more discriminative cross-modal feature space for implicit correspondence learning. In addition, Cross-Modal Query Pruning (CQP) suppresses unreliable candidates during early refinement, improving correspondence refinement for robust pose estimation.

The main contributions of this paper are summarized as follows:

\begin{itemize}
    \item We propose DPA-I2P, a depth-aware implicit correspondence learning framework for robust end-to-end Image-to-Point Cloud Registration. By integrating metric depth and projection-aware visual cues, DPA-I2P constructs a more discriminative cross-modal feature space for correspondence learning.

    \item We introduce RMDE and PVL to enhance cross-modal geometric consistency. RMDE injects metric depth cues into image features to improve their geometric awareness, while PVL transfers projection-consistent visual information to point features.

    \item We propose CQP to suppress unreliable candidates during early correspondence refinement, improving correspondence stability and robustness in pose estimation.

    \item Extensive experiments on KITTI demonstrate that DPA-I2P achieves state-of-the-art registration accuracy, reducing RTE and RRE by 45.0\% and 55.6\%, respectively, over the strongest implicit baseline. Qualitative results on nuScenes further demonstrate its robustness and cross-scene generalization capability.
\end{itemize}
\section{Related Work}

\subsection{Image-to-Point Cloud Registration}

Existing Image-to-Point Cloud (I2P) Registration methods can generally be divided into three categories: explicit correspondence-based methods, correspondence-free methods, and implicit correspondence learning methods.

Explicit correspondence-based methods~\cite{feng20192d3d,pham2020lcd,ren2022corri2p,zhou2023differentiable} first establish 2D--3D correspondences and then estimate camera pose via PnP-RANSAC~\cite{lepetit2009ep,fischler1981random}. To improve robustness, recent works explore stronger feature representations and matching strategies. For example, FreeReg~\cite{wang2024freereg} introduces diffusion-based and depth-enhanced features, while CFI2P~\cite{yao2024quantity} adopts quantity-aware coarse-to-fine matching. Other methods such as DeepI2P~\cite{li2021deepi2p}, EFGHNet~\cite{jeon2022efghnet}, CoFiI2P~\cite{kang2024cofii2p}, and GraphI2P~\cite{bie2025graphi2p} further improve robustness via overlap reasoning or structured matching. However, they remain sensitive to cross-modal feature inconsistency and correspondence quality.

Correspondence-free methods~\cite{li2021deepi2p,yao2024mafreei2p} directly regress camera pose from global cross-modal features without explicit matching. While simplifying the pipeline, they often suffer from unstable performance under large viewpoint changes or weak geometric constraints.

More recently, implicit correspondence learning methods~\cite{li2025implicit,mu2025diff,cheng2025i2p} unify bridge the gap between explicit matching and direct pose regression by learning latent query-based correspondences for end-to-end pose estimation. These methods typically rely on a differentiable probabilistic PnP solver for final pose estimation. However, they still struggle to learn discriminative representations during early correspondence exploration. In contrast, our method constructs a more discriminative feature space by integrating metric depth and projection cues into image and point representations.

\subsection{Monocular Metric Depth Estimation}

Monocular depth estimation provides important geometric priors for 3D vision tasks. Early methods such as MiDaS~\cite{ranftl2020towards} and DPT~\cite{ranftl2021vision} predict relative depth from monocular images, but lack metric-scale consistency, limiting their use in geometry-sensitive tasks such as Image-to-Point Cloud Registration.

Recent methods directly predict metric-scale depth for more reliable geometric reasoning. We adopt UniDepthV2~\cite{piccinelli2025unidepthv2} due to its strong cross-domain generalization ability and accurate metric depth prediction. It also produces pixel-wise confidence, which serves as an indicator of prediction reliability. In our framework, metric depth and confidence cues are incorporated into both image and point representations. Confidence further guides RMDE to emphasize reliable geometry and improves view-consistent vision lifting in PVL, leading to more stable correspondence learning.

\section{Method}
\subsection{Overview}

Given a 3D scene point cloud $P \in \mathbb{R}^{N \times 3}$ and an image $I \in \mathbb{R}^{H \times W \times 3}$ from the same scene, Image-to-Point Cloud Registration aims to estimate the camera pose $T_{gt}=[R_{gt}|t_{gt}]$.
\begin{figure*}[!htbp]
\centering
\includegraphics[width=0.98\textwidth]{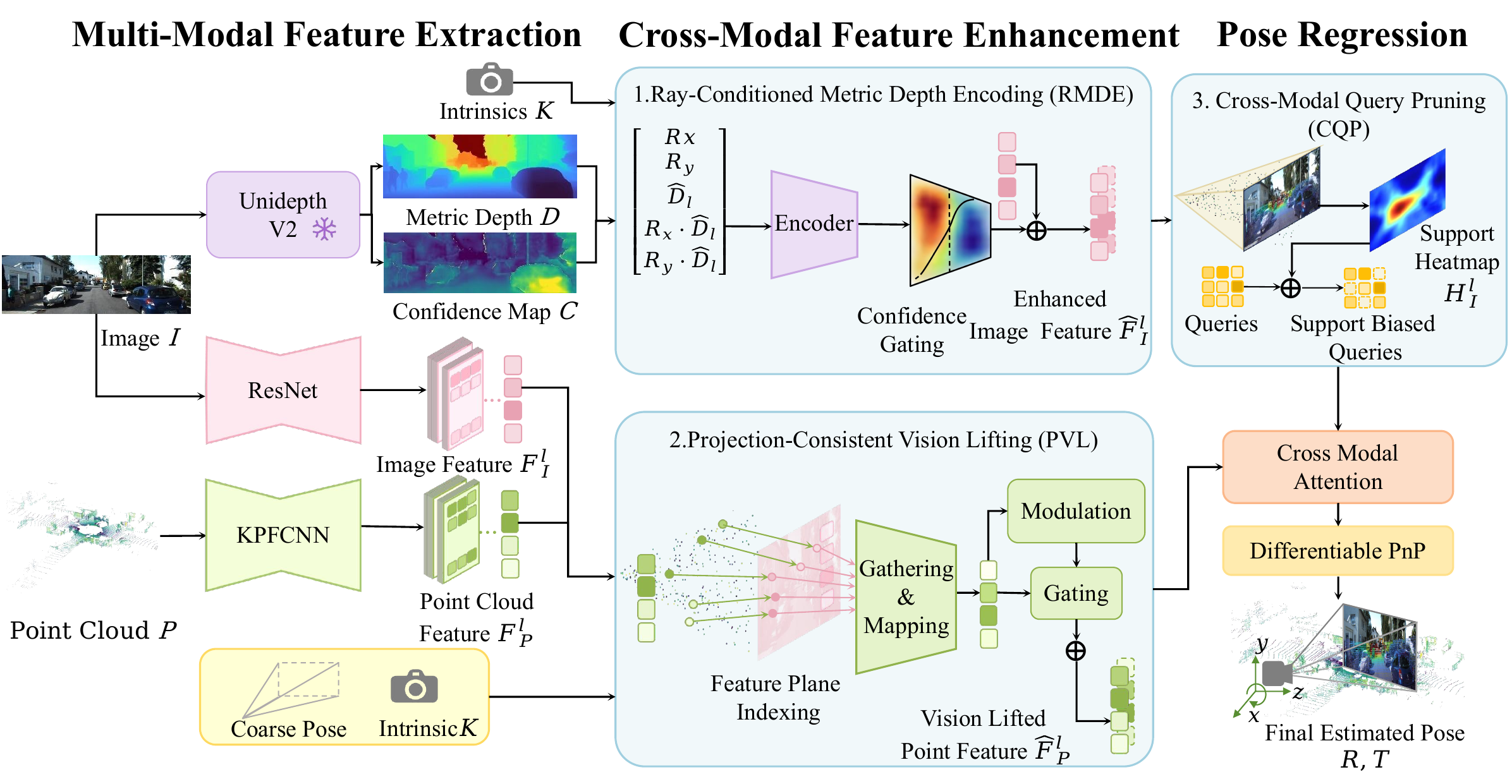}
\caption{Overview of DPA-I2P. Given an image and point cloud, the framework first integrates metric depth cues with a coarse projection-consistent prior via \RMDE{ }and \PVL{ }for feature enhancement, together with an additional \CQP{ }module for query refinement. It then performs implicit correspondence learning for final pose regression.}
\label{fig:overview}
\end{figure*}

An overview of DPA-I2P is illustrated in Fig.~\ref{fig:overview}. Given an image and a point cloud, the framework first extracts multi-scale image and point features and obtains an initial coarse pose estimate~\cite{li2025implicit}.To exploit the monocular depth prior in a structured geometry-aware manner rather than through simple depth concatenation, RMDE constructs uncertainty-aware ray-conditioned depth features to enhance image representations, while PVL injects projection-consistent visual cues into point features under the coarse pose prior. CQP is further introduced during early refinement stages to suppress unreliable correspondence matching. Finally, the refined implicit correspondences are used for camera pose estimation.

\subsection{Ray-Conditioned Metric Depth Encoding}
\label{sec:method_22}

Image features contain strong visual cues but limited metric-scale geometric awareness, which weakens early correspondence reasoning in repetitive or low-texture regions, especially in large-scale outdoor driving environments with sparse LiDAR observations and significant viewpoint variations. To address this issue, we introduce Ray-Conditioned Metric Depth Encoding (RMDE), which converts monocular metric depth into a ray-aware geometric representation before cross-modal correspondence learning.

Given an input image $I \in \mathbb{R}^{H \times W \times 3}$, UniDepthV2 provides a metric depth map $D \in \mathbb{R}^{H \times W}$ and a confidence map $C \in \mathbb{R}^{H \times W}$, both of which are frozen during training. Let $\mathbf{F}_I^l \in \mathbb{R}^{C_l \times H_l \times W_l}$ denote the stage-$l$ image feature. The depth and confidence maps are resized to the same resolution, producing $D^l$ and $C^l$. We normalize the confidence map as $\bar{C}^l=\mathrm{Norm}(C^l)$, where a larger value indicates a more reliable depth prediction.

The camera intrinsics are adapted to the feature resolution:
$f_x^l=f_x\frac{W_l}{W}$,
$f_y^l=f_y\frac{H_l}{H}$,
$c_x^l=c_x\frac{W_l}{W}$, and
$c_y^l=c_y\frac{H_l}{H}$.
For each feature location $i=(u_i^l,v_i^l)$, we compute the camera-ray coordinate as
\begin{equation}
r_{x,i}^l = \frac{u_i^l - c_x^l}{f_x^l},
\quad
r_{y,i}^l = \frac{v_i^l - c_y^l}{f_y^l},
\end{equation}
and define the corresponding ray direction as
\begin{equation}
\tilde{\mathbf{d}}_i^l =
\left[
r_{x,i}^l,\;
r_{y,i}^l,\;
1
\right]^\top,
\quad
\mathbf{d}_i^l =
\frac{\tilde{\mathbf{d}}_i^l}
{\|\tilde{\mathbf{d}}_i^l\|_2}.
\end{equation}

Instead of directly concatenating the depth map with image features, RMDE samples $N$ candidate points along each viewing ray using the predicted metric depth as the reference surface. Specifically, we define an uncertainty-aware sampling radius
\begin{equation}
s_i^l =
s_{\min}^l +
(s_{\max}^l-s_{\min}^l)
(1-\bar{C}_i^l),
\end{equation}
where $s_{\min}^l$ and $s_{\max}^l$ are stage-specific positive constants. A higher confidence produces a narrower sampling range, while a lower confidence allows a broader ray interval. Given a set of fixed offsets $\{\rho_k\}_{k=1}^{N}$ with $\rho_k \in [-1,1]$, the sampled depth values are
\begin{equation}
z_{i,k}^l =
\max
\left(
D_i^l + \rho_k s_i^l,\;
\epsilon
\right),
\quad k=1,\ldots,N.
\end{equation}
The corresponding 3D point in the camera coordinate system is then obtained by
\begin{equation}
\mathbf{p}_{i,k}^l =
z_{i,k}^l
\tilde{\mathbf{d}}_i^l .
\end{equation}

For each sampled point, we encode its 3D location, ray direction, and depth offset with respect to the predicted surface. The normalized depth offset is defined as
\begin{equation}
\Delta z_{i,k}^l =
\log(z_{i,k}^l+\epsilon)
-
\log(D_i^l+\epsilon).
\end{equation}
The ray-conditioned geometric descriptor of the $k$-th sampled point is
\begin{equation}
\mathbf{h}_{i,k}^l =
\Big[
\gamma_p(\mathbf{p}_{i,k}^l);
\gamma_d(\mathbf{d}_i^l);
\gamma_z(\Delta z_{i,k}^l);
\bar{C}_i^l
\Big],
\end{equation}
where $\gamma_p(\cdot)$, $\gamma_d(\cdot)$, and $\gamma_z(\cdot)$ denote lightweight positional encodings for 3D location, ray direction, and depth offset, respectively. A shared encoder maps the descriptor to a sampled ray feature:
\begin{equation}
\mathbf{e}_{i,k}^l =
\phi^l(\mathbf{h}_{i,k}^l),
\quad
\mathbf{e}_{i,k}^l \in \mathbb{R}^{C_l}.
\end{equation}

To aggregate sampled features along the ray, we assign larger weights to samples closer to the predicted metric depth and make the weighting less peaked when the depth confidence is low:
\begin{equation}
\tilde{w}_{i,k}^l =
\bar{C}_i^l
\exp
\left(
-\frac{
|\Delta z_{i,k}^l|
}{
\tau_l+\epsilon
}
\right)
+
(1-\bar{C}_i^l),
\end{equation}
\begin{equation}
w_{i,k}^l =
\frac{
\tilde{w}_{i,k}^l
}{
\sum_{j=1}^{N}
\tilde{w}_{i,j}^l
}.
\end{equation}
The ray-conditioned metric depth encoding for location $i$ is obtained by a confidence-weighted aggregation:
\begin{equation}
\mathbf{e}_{i}^l =
\sum_{k=1}^{N}
w_{i,k}^l
\mathbf{e}_{i,k}^l .
\end{equation}

Finally, RMDE injects the aggregated ray-aware geometric representation into the image feature through a gated residual update. Let $\mathbf{f}_{I,i}^l$ denote the image feature at location $i$. The residual gate is computed as
\begin{equation}
\mathbf{g}_{i}^l =
\sigma
\left(
\eta^l
\left(
[
\mathbf{f}_{I,i}^l;
\mathbf{e}_{i}^l;
\bar{C}_i^l
]
\right)
\right),
\end{equation}
where $\eta^l(\cdot)$ is a lightweight projection layer. The enhanced image feature is
\begin{equation}
\hat{\mathbf{f}}_{I,i}^l
=
\mathbf{f}_{I,i}^l
+
\bar{C}_i^l
\left(
\mathbf{g}_{i}^l
\odot
\mathbf{e}_{i}^l
\right).
\end{equation}
The enhanced image feature map is then reconstructed as
\begin{equation}
\hat{\mathbf{F}}_I^l
=
\mathrm{Reshape}
\left(
\{
\hat{\mathbf{f}}_{I,i}^l
\}_{i=1}^{H_lW_l}
\right).
\end{equation}

In this way, RMDE does not merely append monocular depth to image features. Instead, it represents each image token as an uncertainty-aware local ray distribution around the predicted metric surface, allowing image features to carry camera-aware geometric cues before cross-modal correspondence learning.

\subsection{Projection-Consistent Vision Lifting}
\label{sec:method_23}

Although RMDE enhances image-side geometry, point features remain view-independent and lack vision consistency with the image branch. We therefore project 3D points onto the image feature plane using the coarse pose prior and lift projection-consistent vision cues into point features.

Let $\mathbf{X}_I^l=\mathrm{Flatten}(\hat{\mathbf{F}}_I^l)\in\mathbb{R}^{H_lW_l \times C_I^l}$ denote the flattened image tokens, and let $\mathbf{F}_P^l \in \mathbb{R}^{N_l \times C_P^l}$ denote the stage-$l$ point features. Given the coarse pose $\mathbf{T}_f=(\mathbf{R}_f,\mathbf{t}_f)$ and camera intrinsics $\mathbf{K}$, each point $\mathbf{p}_i^l$ is projected onto the stage-$l$ image feature plane as
\begin{equation}
\bar{\mathbf{u}}_i^l
=
\lfloor
\Gamma^l(
\pi(\mathbf{p}_i^l;\mathbf{K},\mathbf{R}_f,\mathbf{t}_f)
)
\rceil,
\end{equation}
with validity indicator $m_i^l \in \{0,1\}$. The corresponding image feature is gathered by $\mathbf{z}_i^l=\mathcal{G}(\mathbf{X}_I^l,\bar{\mathbf{u}}_i^l)$, where $\mathcal{G}(\cdot)$ denotes center-cell feature gathering, and projected into the point feature space as $\bar{\mathbf{z}}_i^l=\psi^l(\mathbf{z}_i^l)$.

To reduce sensitivity to coarse pose errors, the vision cue is incorporated through a bounded residual update:
\begin{equation}
\hat{\mathbf{f}}_{P,i}^l
=
\mathbf{f}_{P,i}^l
+
\delta^l m_i^l
\big(
\tanh(\mathbf{w}^l)
\odot
\bar{\mathbf{z}}_i^l
\big),
\end{equation}
where $\mathbf{w}^l \in \mathbb{R}^{C_P^l}$ is a learnable channel-wise modulation vector and $\delta^l \in \{0,1\}$ is a stage indicator. PVL is enabled only in early stages and disabled during late refinement stages to avoid noisy vision priors. Invalid projections ($m_i^l=0$) preserve the original point features. The enhanced point features are
\begin{equation}
\hat{\mathbf{F}}_P^l
=
\{
\hat{\mathbf{f}}_{P,i}^l
\}_{i=1}^{N_l}.
\end{equation}

\subsection{Cross-Modal Query Pruning}

Although RMDE and PVL improve cross-modal representations, early correspondence queries may still drift toward geometrically unsupported regions. To stabilize correspondence exploration, we introduce a projective support prior.

Let $T^{l-1}=(R^{l-1},t^{l-1})$ denote the pose estimate before the $l$-th refinement stage. Given the stage-wise point set $\{\mathbf{p}_i^l\}_{i=1}^{N_l}$, each point is projected onto the stage-$l$ feature plane as
\begin{equation}
\hat{\mathbf{u}}_i^l
=
\Gamma^l(
\pi(\mathbf{p}_i^l;K,R^{l-1},t^{l-1})
),
\end{equation}
with validity indicator $m_i^l \in \{0,1\}$. Based on the projected points, we construct a dense support prior:
\begin{equation}
\mathcal{S}^l(x)
=
\sum_{i=1}^{N_l}
m_i^l
\exp
\left(
-\frac{
\|x-\hat{\mathbf{u}}_i^l\|_2^2
}{
2\sigma_l^2
}
\right).
\end{equation}

After normalization, the support prior is flattened into $\mathbf{s}^l=\mathrm{Flatten}(\mathcal{S}^l)$. Using the flattened image tokens $\mathbf{X}_I^l \in \mathbb{R}^{H_lW_l \times C_l}$ and image-side queries $Q_{\mathrm{img}}^l \in \mathbb{R}^{N_q \times C_l}$, we compute the support-aware query-token similarity:
\begin{equation}
B_I^l
=
\frac{
Q_{\mathrm{img}}^l
(\mathbf{X}_I^l)^\top
}{
\sqrt{C_l}
}
+
\beta_l \log(\mathbf{s}^l+\epsilon),
\end{equation}
where $\beta_l$ is a stage-dependent weight and $\epsilon$ avoids numerical instability.

The image-side heatmap is then computed as $H_I^l=\mathrm{Softmax}(B_I^l)$, and the corresponding 2D keypoints are obtained by $K_I^l = H_I^lE_I^l$, where $E_I^l \in \mathbb{R}^{H_lW_l \times 2}$ denotes the stage-wise image coordinate matrix.

For regions with extremely low support, we further construct a conservative pruning mask $M_{\mathrm{sup}}^l = \mathbb{I}(\mathbf{s}^l<\tau_l)$, where $\tau_l$ is a small threshold. Query pruning is applied only during early refinement stages and gradually relaxed later.

\subsection{Loss Function}

The metric depth network remains frozen during training. The overall objective jointly optimizes pose regression and support-guided query regularization.

Let the ground-truth pose be $T_{\mathrm{gt}}=(R_{\mathrm{gt}},t_{\mathrm{gt}})$, the coarse pose be $T_f=(R_f,t_f)$, and the refined pose at stage $l$ be $\hat{T}^l=(\hat{R}^l,\hat{t}^l)$. The pose regression loss is defined as
\begin{equation}
\mathcal{L}_{\mathrm{pose}}(T,T_{\mathrm{gt}})
=
\|\hat{t}^l-t_{\mathrm{gt}}\|_1
+
\lambda_R
d_R(\hat{R}^l,R_{\mathrm{gt}}),
\end{equation}
where $d_R(\cdot,\cdot)$ denotes the geodesic rotation distance and $\lambda_R$ balances translation and rotation terms.

To align early correspondence exploration with projective support, we introduce a support-guided regularization loss:
\begin{equation}
\mathcal{L}_{\mathrm{sup}}
=
\sum_{l\in\mathcal{E}}
\frac{1}{N_q}
\sum_{q=1}^{N_q}
\mathrm{KL}
\left(
H_{I,q}^l
\,\|\,
\tilde{\mathbf{s}}_q^l
\right).
\end{equation}
where $\mathcal{E}$ denotes the set of early refinement stages, and $\tilde{\mathbf{s}}_q^l$ denotes the normalized projective support distribution for query $q$.

The overall objective is
\begin{equation}
\mathcal{L}
=
\lambda_c
\mathcal{L}_{\mathrm{pose}}(T_c,T_{\mathrm{gt}})
+
\lambda_p
\sum_{l=1}^{L}
\omega_l
\mathcal{L}_{\mathrm{pose}}(T,T_{\mathrm{gt}})
+
\lambda_s
\mathcal{L}_{\mathrm{sup}},
\end{equation}
where $\omega_l$ denotes the stage weight and $\lambda_c,\lambda_p,\lambda_s$ balance different loss terms. Following~\cite{li2025implicit}, the coarse pose branch is first optimized independently, after which the entire framework is trained end-to-end.

\section{Experiments}
In this section, we first introduce the implementation details. Then, we present quantitative results on the KITTI dataset and qualitative visualizations on both KITTI and nuScenes datasets. Finally, we conduct a series of ablation studies to verify the effectiveness of each component.

\subsection{Implementation Details}
Our framework employs a 4-stage ResNet~\cite{he2016deep} with FPN~\cite{lin2017feature} as the image backbone and a 4-stage KPFCNN~\cite{thomas2019kpconv} as the point backbone, both outputting 128-dimensional features. We utilize 4-head attention~\cite{vaswani2017attention} across all layers, with $N_q=128$ correspondence queries and a refinement depth of $L=3$. metric depth and confidence maps are pre-computed offline using UniDepthV2 and kept frozen to provide geometric priors for correspondence learning, without fine-tuning the depth estimator. Specifically, \RMDE{} is applied at all four scales, while \PVL{} and \CQP{} are utilized only in the first two stages. The model is trained for 40 epochs using the Adam optimizer (batch size 4, initial learning rate $2\times10^{-4}$, weight decay $10^{-6}$) on a single NVIDIA RTX 4090 GPU.

\subsection{Datasets and Evaluation Metrics}
\subsubsection{KITTI~\cite{geiger2012kitti}.}
We evaluate on the KITTI dataset, comprising 22 synchronized image--LiDAR sequences, and 11 sequences provide ground-truth calibration files. Following the standard protocol~\cite{zhou2023differentiable}, sequences 0--8 are used for training and sequences 9--10 for evaluation. Misregistration transformations are generated using a 2D translation on the ground plane within $\pm 10$~m and a rotation around the up-axis with no limited range. Input images are resized to $160 \times 512$, and point clouds are downsampled to 40,960 points.
\subsubsection{nuScenes~\cite{caesar2020nuscenes}.}
We further conduct qualitative evaluation on the nuScenes benchmark using the official 150 test scenes. Image and point cloud pairs are obtained via the nuScenes SDK, where point clouds are accumulated from adjacent frames to ensure sufficient density. Input images are resized to $160\times320$, and point clouds are uniformly downsampled to 40,960 points. 
\subsubsection{Evaluation Metrics.}
We adopt average Relative Translational Error (RTE), average Relative Rotation Error (RRE), and registration accuracy (Acc). Acc is defined as the proportion of registrations satisfying $\mathrm{RTE} < 2~\mathrm{m}$ and $\mathrm{RRE} < 5^{\circ}$.

\subsection{Registration Accuracy}
\subsubsection{Quantitative Comparison.}
We report the quantitative results in Table~\ref{tab:main_results}. DPA-I2P achieves the best performance across all evaluation metrics on the KITTI dataset. Compared with ICLI2P~\cite{li2025implicit}, our method reduces RTE from 0.20~m to 0.11~m and RRE from 1.24$^\circ$ to 0.55$^\circ$, while achieving the highest registration accuracy of 99.70\%. We attribute the superior performance to three factors. RMDE reduces ambiguous image matching by introducing metric depth and ray-direction cues. PVL improves cross-modal feature compatibility by lifting projection-consistent vision cues to point features. CQP further stabilizes early correspondence refinement by suppressing projectively unsupported matching.

\begin{table*}[!htbp]
\caption{Registration accuracy on the KITTI and nuScenes datasets. Lower is better for RTE and RRE, higher is better for Acc.}
\label{tab:main_results}
\centering
\renewcommand{\arraystretch}{1.03}
\setlength{\tabcolsep}{2.4pt}
\begin{tabular}{l|ccc|ccc}
\hline
\multirow{2}{*}{Methods}
& \multicolumn{3}{c|}{KITTI}
& \multicolumn{3}{c}{nuScenes} \\
\cline{2-7}
& RTE$\downarrow$
& RRE$\downarrow$
& Acc.$\uparrow$
& RTE$\downarrow$
& RRE$\downarrow$
& Acc.$\uparrow$ \\
\hline
Grid Cls.+PnP~\cite{li2021deepi2p}
& 3.64$\pm$3.46 & 19.19$\pm$28.96 & 11.22
& 3.02$\pm$2.40 & 12.66$\pm$21.01 & 2.45 \\
DeepI2P(3D)~\cite{li2021deepi2p}
& 4.06$\pm$3.54 & 24.73$\pm$31.69 & 3.77
& 2.88$\pm$2.12 & 20.65$\pm$12.24 & 2.26 \\
DeepI2P(2D)~\cite{li2021deepi2p}
& 3.59$\pm$3.21 & 11.66$\pm$18.16 & 25.95
& 2.78$\pm$1.99 & 4.80$\pm$6.21 & 38.10 \\
CorrI2P~\cite{ren2022corri2p}
& 3.78$\pm$65.16 & 5.89$\pm$20.34 & 72.42
& 3.04$\pm$60.76 & 3.73$\pm$9.03 & 49.00 \\
VP2P-Match~\cite{zhou2023differentiable}
& 0.75$\pm$1.13 & 3.29$\pm$7.99 & 83.04
& 0.89$\pm$1.44 & 2.15$\pm$7.03 & 88.33 \\
ICLI2P~\cite{li2025implicit}
& 0.20$\pm$0.21 & 1.24$\pm$2.34 & 97.49
& 0.63$\pm$0.44 & 2.13$\pm$3.75 & 90.94 \\
\hline
DPA-I2P
& \textbf{0.11$\pm$0.12} & \textbf{0.55$\pm$0.67} & \textbf{99.70}
& \textbf{0.54$\pm$0.37} & \textbf{1.92$\pm$3.81} & \textbf{92.02} \\
\hline
\end{tabular}
\end{table*}
\newcommand{\figann}[1]{\footnotesize #1}
\newcommand{\figcol}[1]{\footnotesize\textbf{#1}}
\newcommand{\figmetricann}{\footnotesize RTE(m)/RRE($^\circ$)}

\newcommand{\qualcolw}{0.19\textwidth}
\newcommand{\qualimgw}{0.88\linewidth}
\newcommand{\qualgap}{0.8mm}

\begin{figure*}[!t]
\centering
\begin{minipage}[t]{\qualcolw}\centering
\figcol{\mbox{VP2P-Match}}
\end{minipage}\hspace{\qualgap}
\begin{minipage}[t]{\qualcolw}\centering
\figcol{\mbox{ICLI2P}}
\end{minipage}\hspace{\qualgap}
\begin{minipage}[t]{\qualcolw}\centering
\figcol{Ours}
\end{minipage}\hspace{\qualgap}
\begin{minipage}[t]{\qualcolw}\centering
\figcol{\mbox{Ground Truth}}
\end{minipage}\\[0.8mm]

\begin{minipage}[t]{\qualcolw}\centering
\includegraphics[width=\qualimgw]{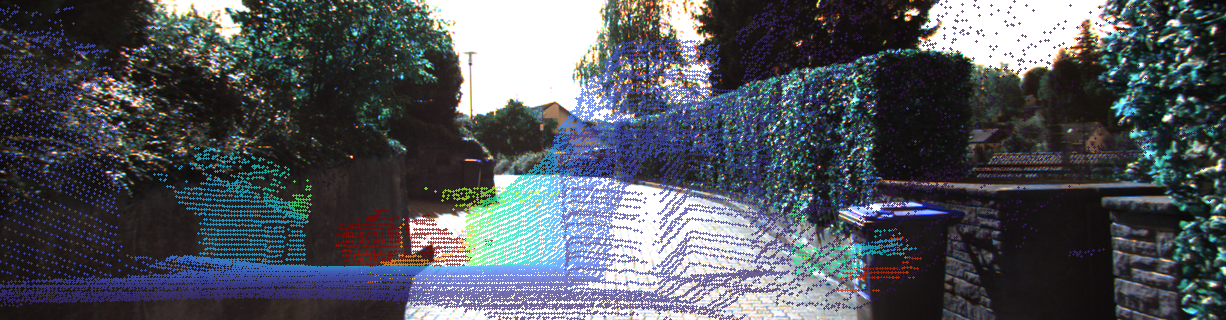}
\figann{$2.47\,\mathrm{m}/12.93^\circ$}
\end{minipage}\hspace{\qualgap}
\begin{minipage}[t]{\qualcolw}\centering
\includegraphics[width=\qualimgw]{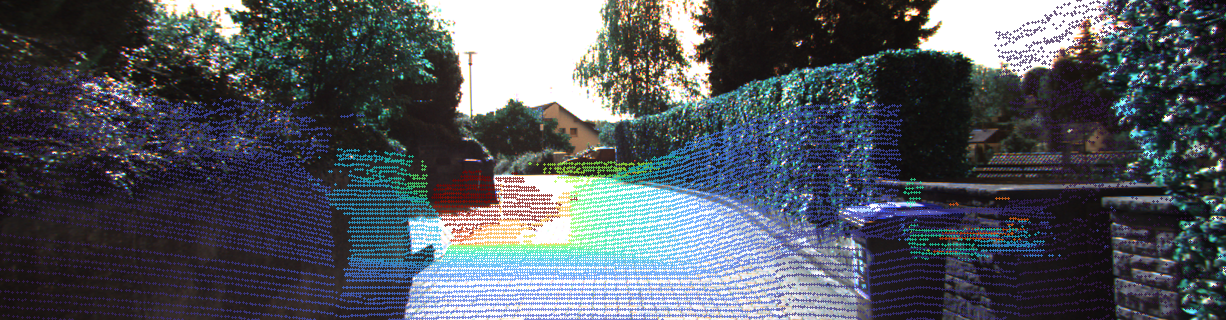}
\figann{$0.22\,\mathrm{m}/0.14^\circ$}
\end{minipage}\hspace{\qualgap}
\begin{minipage}[t]{\qualcolw}\centering
\includegraphics[width=\qualimgw]{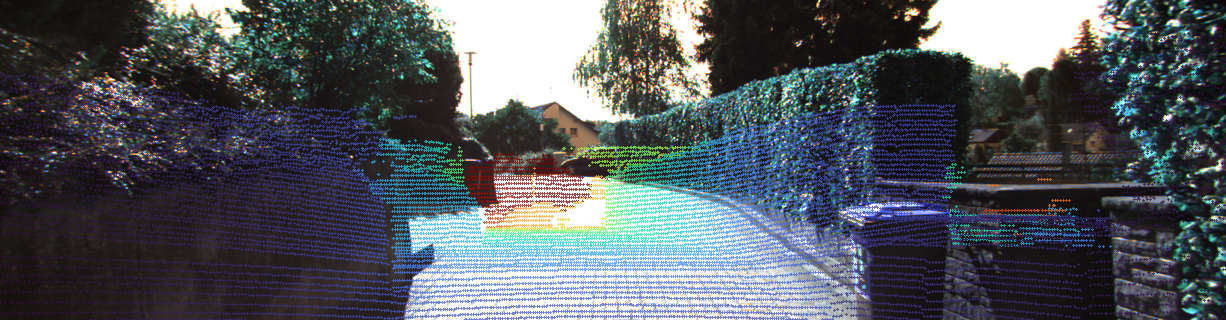}
\figann{$0.09\,\mathrm{m}/0.12^\circ$}
\end{minipage}\hspace{\qualgap}
\begin{minipage}[t]{\qualcolw}\centering
\includegraphics[width=\qualimgw]{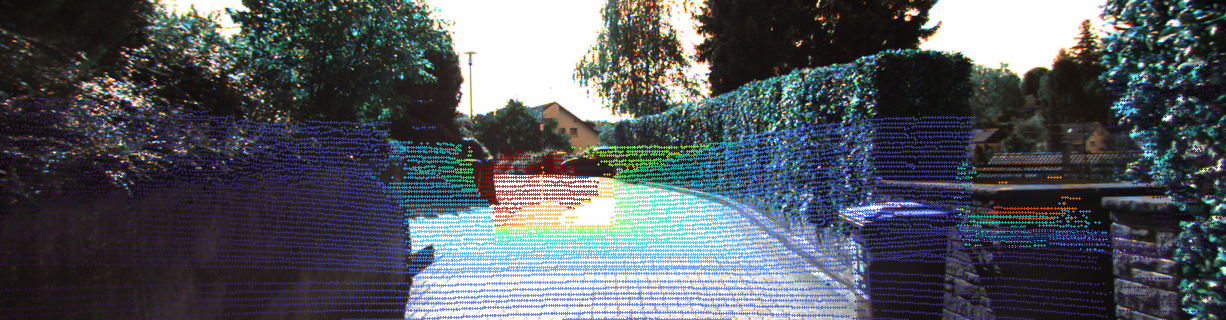}
\figmetricann
\end{minipage}\\[1mm]

\begin{minipage}[t]{\qualcolw}\centering
\includegraphics[width=\qualimgw]{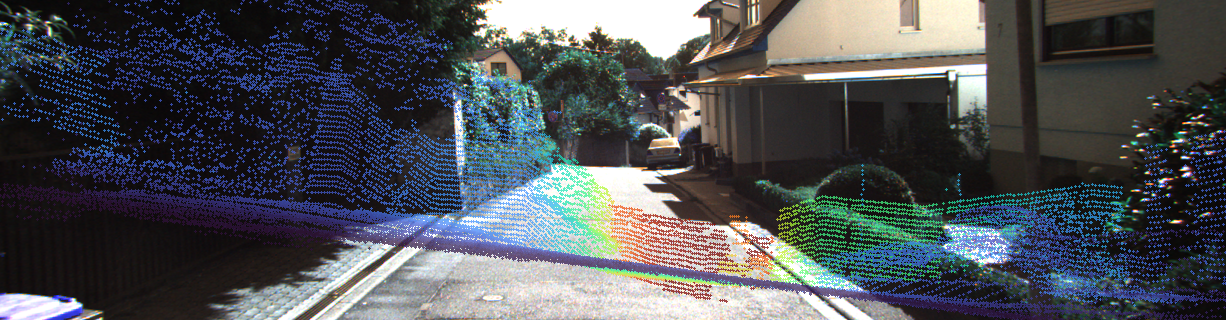}
\figann{$1.98\,\mathrm{m}/11.37^\circ$}
\end{minipage}\hspace{\qualgap}
\begin{minipage}[t]{\qualcolw}\centering
\includegraphics[width=\qualimgw]{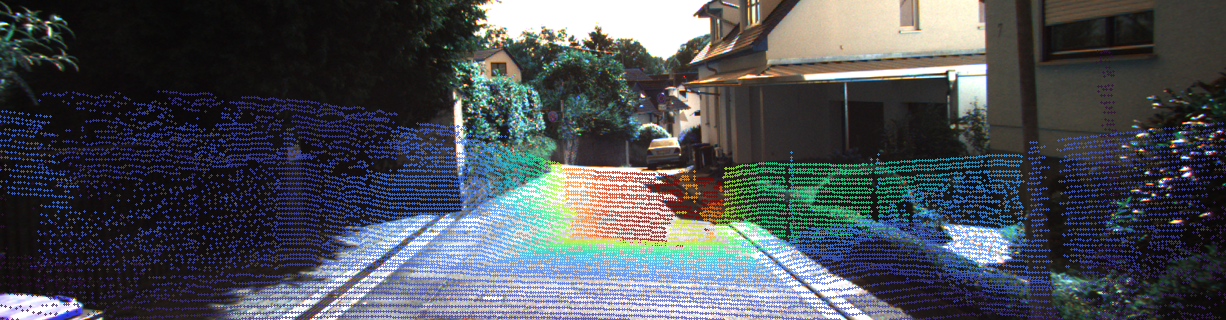}
\figann{$0.27\,\mathrm{m}/0.53^\circ$}
\end{minipage}\hspace{\qualgap}
\begin{minipage}[t]{\qualcolw}\centering
\includegraphics[width=\qualimgw]{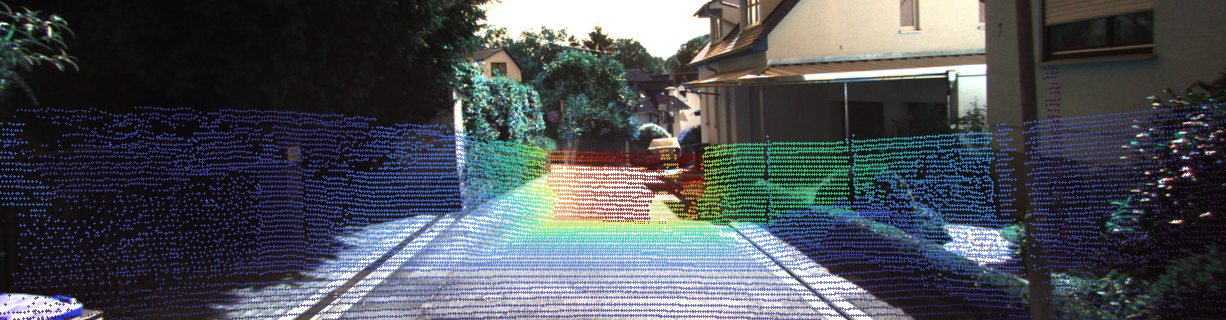}
\figann{$0.11\,\mathrm{m}/0.42^\circ$}
\end{minipage}\hspace{\qualgap}
\begin{minipage}[t]{\qualcolw}\centering
\includegraphics[width=\qualimgw]{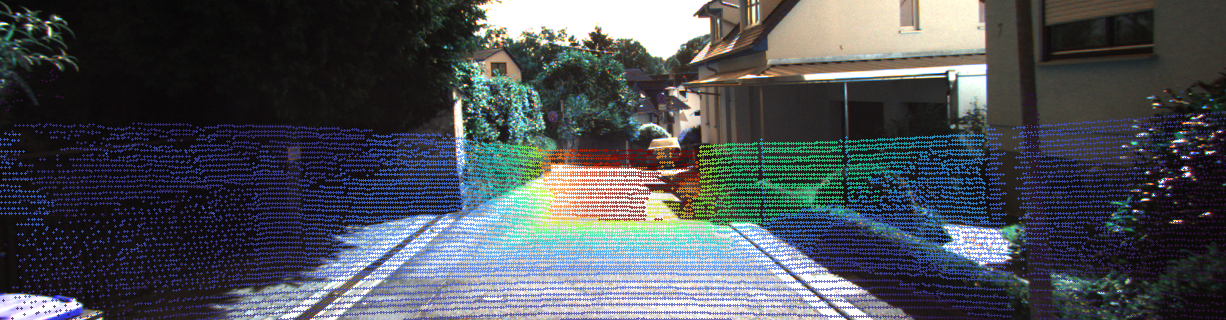}
\figmetricann
\end{minipage}\\[1mm]

\begin{minipage}[t]{\qualcolw}\centering
\includegraphics[width=\qualimgw]{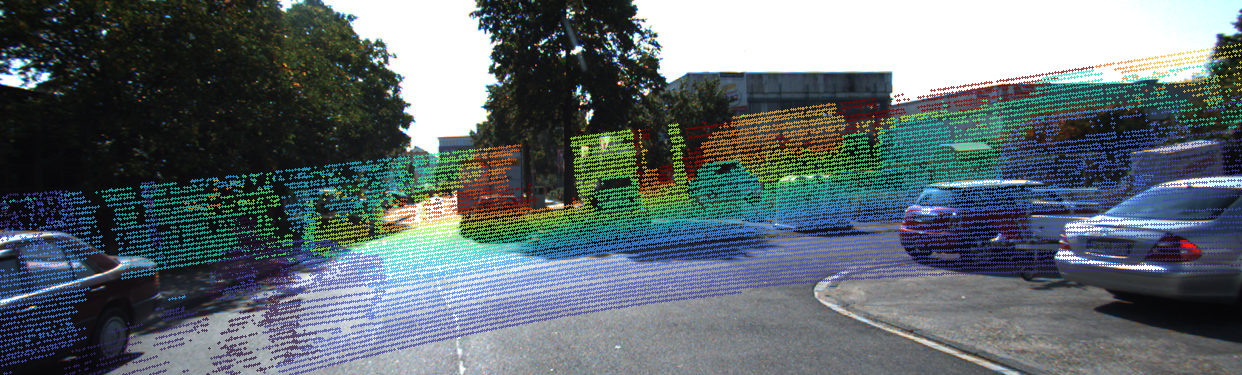}
\figann{$1.87\,\mathrm{m}/10.43^\circ$}
\end{minipage}\hspace{\qualgap}
\begin{minipage}[t]{\qualcolw}\centering
\includegraphics[width=\qualimgw]{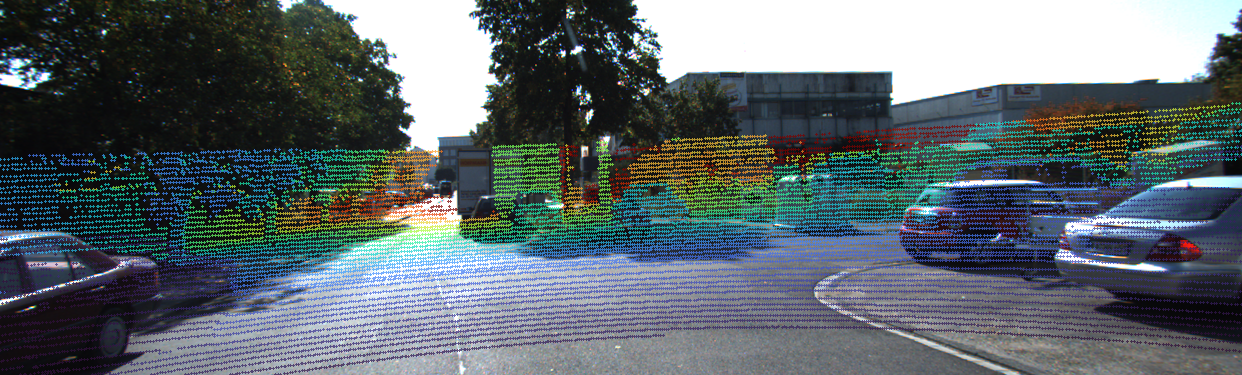}
\figann{$0.37\,\mathrm{m}/2.85^\circ$}
\end{minipage}\hspace{\qualgap}
\begin{minipage}[t]{\qualcolw}\centering
\includegraphics[width=\qualimgw]{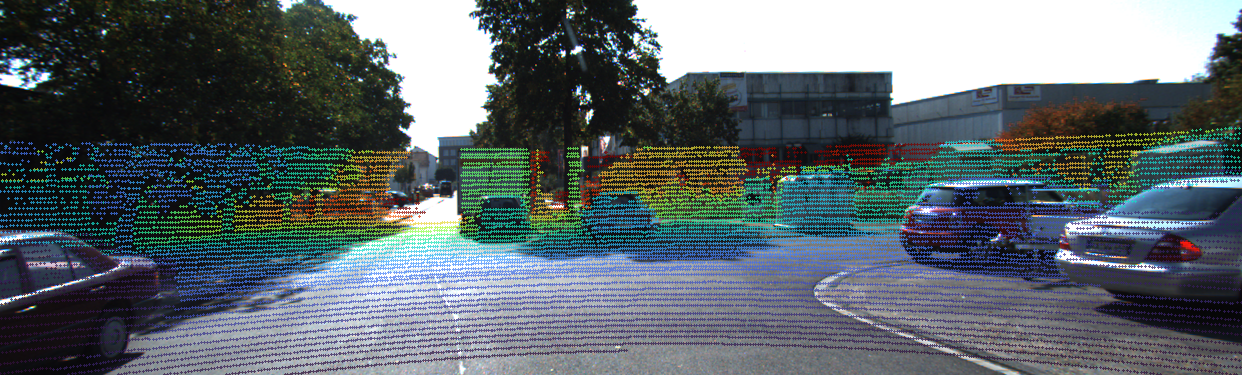}
\figann{$0.14\,\mathrm{m}/0.97^\circ$}
\end{minipage}\hspace{\qualgap}
\begin{minipage}[t]{\qualcolw}\centering
\includegraphics[width=\qualimgw]{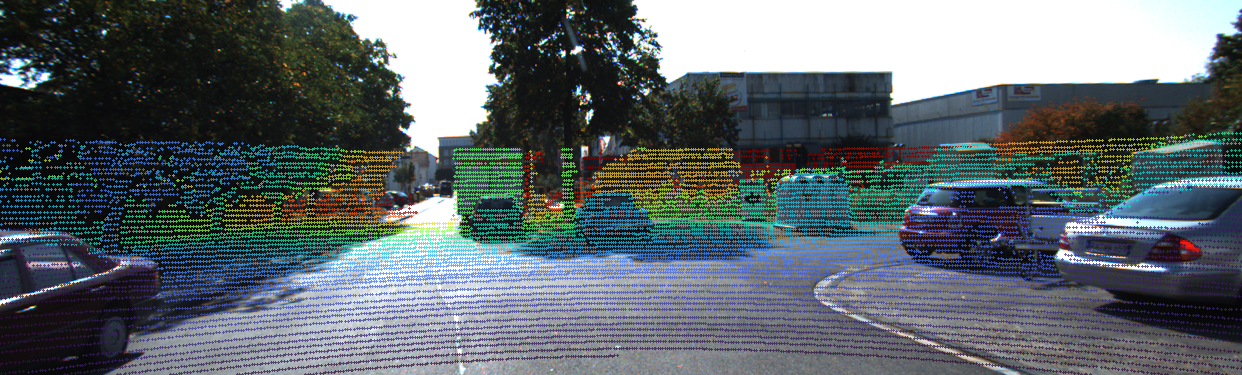}
\figmetricann
\end{minipage}

\caption{Qualitative comparison of image-to-point cloud registration results on the KITTI dataset.}
\label{fig:kitti_qualitative}
\end{figure*}

\begin{figure*}[!t]
\centering
\begin{minipage}[t]{\qualcolw}\centering
\figcol{\mbox{VP2P-Match}}
\end{minipage}\hspace{\qualgap}
\begin{minipage}[t]{\qualcolw}\centering
\figcol{\mbox{ICLI2P}}
\end{minipage}\hspace{\qualgap}
\begin{minipage}[t]{\qualcolw}\centering
\figcol{Ours}
\end{minipage}\hspace{\qualgap}
\begin{minipage}[t]{\qualcolw}\centering
\figcol{\mbox{Ground Truth}}
\end{minipage}\\[0.8mm]

\begin{minipage}[t]{\qualcolw}\centering
\includegraphics[width=\qualimgw]{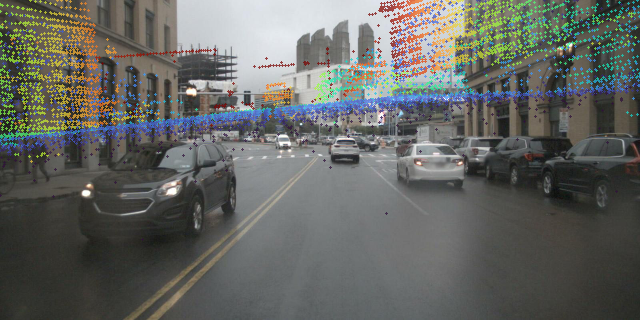}
\figann{$2.23\,\mathrm{m}/9.16^\circ$}
\end{minipage}\hspace{\qualgap}
\begin{minipage}[t]{\qualcolw}\centering
\includegraphics[width=\qualimgw]{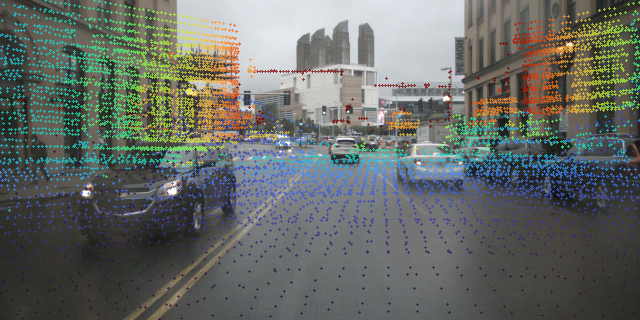}
\figann{$1.03\,\mathrm{m}/4.82^\circ$}
\end{minipage}\hspace{\qualgap}
\begin{minipage}[t]{\qualcolw}\centering
\includegraphics[width=\qualimgw]{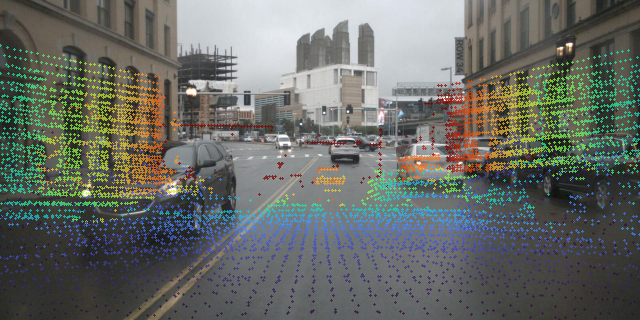}
\figann{$0.85\,\mathrm{m}/3.18^\circ$}
\end{minipage}\hspace{\qualgap}
\begin{minipage}[t]{\qualcolw}\centering
\includegraphics[width=\qualimgw]{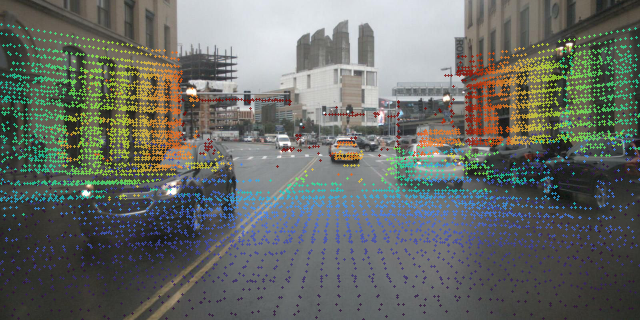}
\figmetricann
\end{minipage}\\[1mm]

\begin{minipage}[t]{\qualcolw}\centering
\includegraphics[width=\qualimgw]{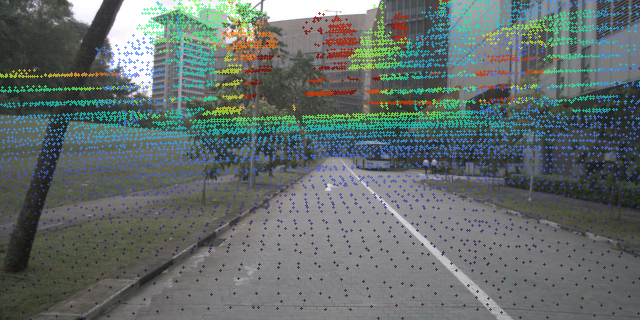}
\figann{$1.75\,\mathrm{m}/8.49^\circ$}
\end{minipage}\hspace{\qualgap}
\begin{minipage}[t]{\qualcolw}\centering
\includegraphics[width=\qualimgw]{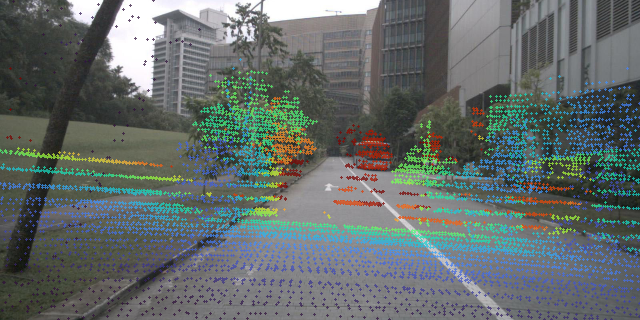}
\figann{$0.87\,\mathrm{m}/5.32^\circ$}
\end{minipage}\hspace{\qualgap}
\begin{minipage}[t]{\qualcolw}\centering
\includegraphics[width=\qualimgw]{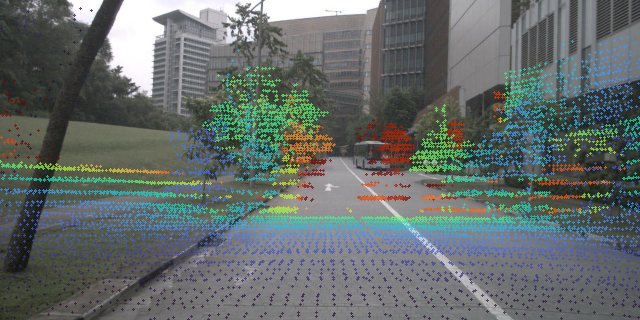}
\figann{$0.69\,\mathrm{m}/3.14^\circ$}
\end{minipage}\hspace{\qualgap}
\begin{minipage}[t]{\qualcolw}\centering
\includegraphics[width=\qualimgw]{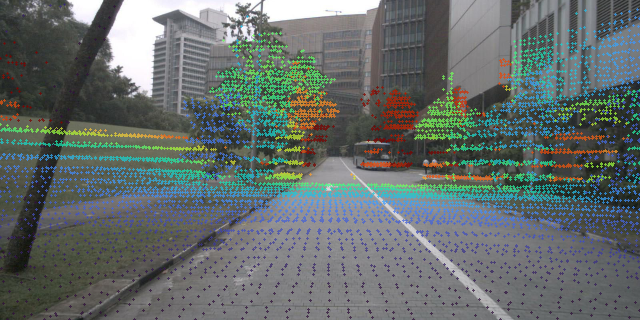}
\figmetricann
\end{minipage}\\[1mm]

\begin{minipage}[t]{\qualcolw}\centering
\includegraphics[width=\qualimgw]{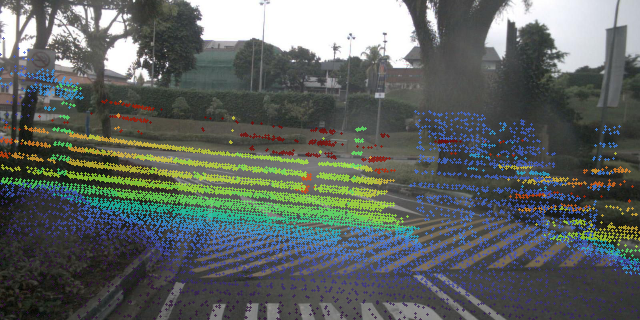}
\figann{$1.45\,\mathrm{m}/5.46^\circ$}
\end{minipage}\hspace{\qualgap}
\begin{minipage}[t]{\qualcolw}\centering
\includegraphics[width=\qualimgw]{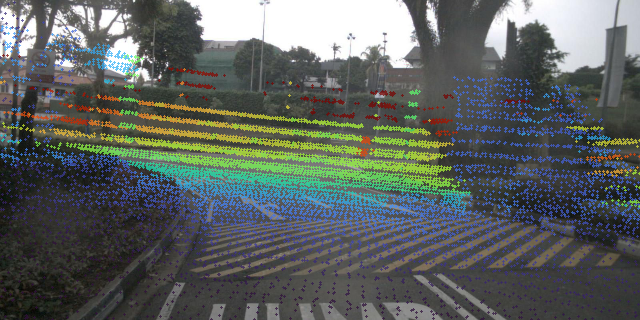}
\figann{$0.57\,\mathrm{m}/2.29^\circ$}
\end{minipage}\hspace{\qualgap}
\begin{minipage}[t]{\qualcolw}\centering
\includegraphics[width=\qualimgw]{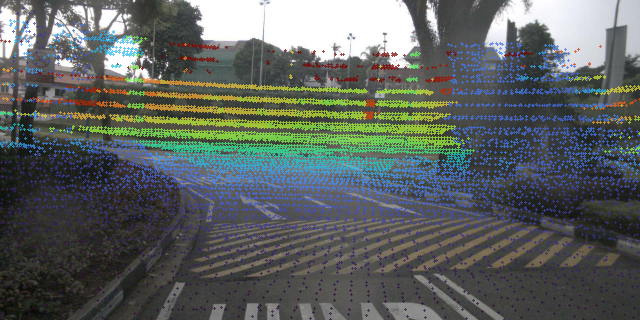}
\figann{$0.28\,\mathrm{m}/1.69^\circ$}
\end{minipage}\hspace{\qualgap}
\begin{minipage}[t]{\qualcolw}\centering
\includegraphics[width=\qualimgw]{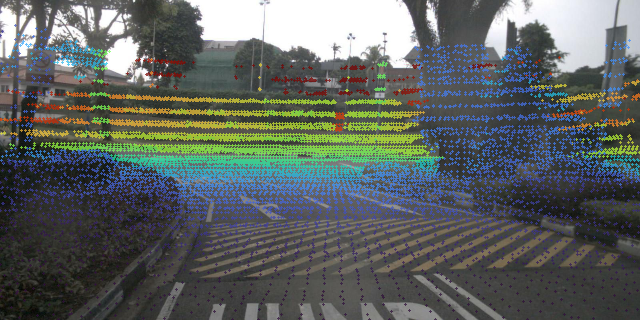}
\figmetricann
\end{minipage}

\caption{Qualitative comparison of image-to-point cloud registration results on the nuScenes dataset.}
\label{fig:nus_qualitative}
\end{figure*}

\subsubsection{Qualitative Comparison.}
\begin{figure}[!t]
\centering
\begin{tabular}{@{}cc@{}}
\includegraphics[width=0.44\linewidth]{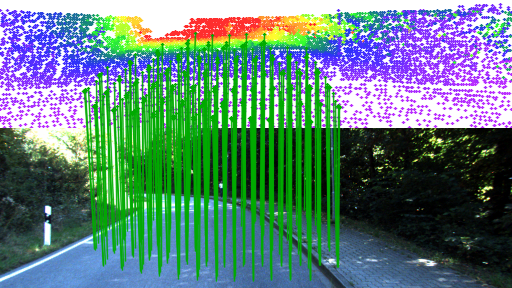} &
\includegraphics[width=0.44\linewidth]{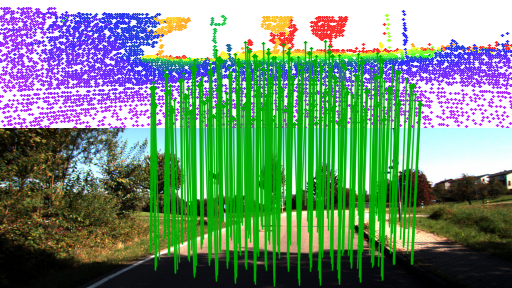} \\
\footnotesize 0.09m / 0.45$^\circ$ & \footnotesize 0.08m / 0.37$^\circ$ \\[1.5ex]
\includegraphics[width=0.44\linewidth]{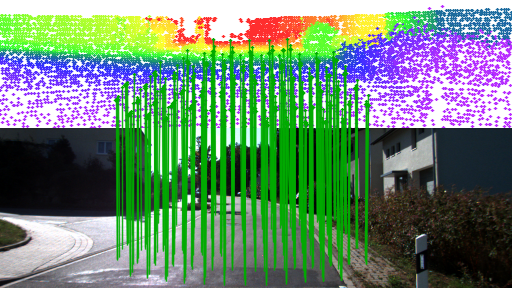} &
\includegraphics[width=0.44\linewidth]{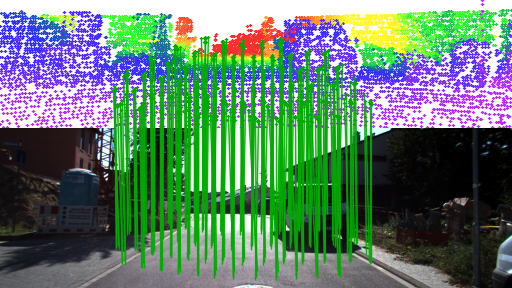} \\
\footnotesize 0.04m / 0.61$^\circ$ & \footnotesize 0.08m / 0.26$^\circ$
\end{tabular}
\caption{Visual illustration of 2D-3D correspondences and registration accuracy. Green lines represent correct correspondences.}
\label{fig:vis_correspondence}
\end{figure}

Qualitative comparisons on KITTI and nuScenes are shown in Fig.~\ref{fig:kitti_qualitative} and Fig.~\ref{fig:nus_qualitative}, respectively. For visualization, we project the point clouds into the image space using the predicted poses from different methods and the camera intrinsics, where colors encode depth. Compared with VP2P-Match~\cite{zhou2023differentiable} and ICLI2P~\cite{li2025implicit}, our method achieves more accurate registration results in challenging driving scenes. We further provide 2D--3D correspondence visualizations in Fig.~\ref{fig:vis_correspondence}. The projected points of our method exhibit more accurate alignment with image structures, indicating that the proposed implicit correspondence learning framework can establish more reliable 2D--3D correspondences.

\subsection{Ablation Study}

In this section, we conduct ablation studies on the KITTI dataset to validate the effectiveness of each component and analyze several key design choices in our framework. 

We evaluate three variants in Table~\ref{tab:ablation_modules}: (1) w/o \RMDE{} removes ray-conditioned metric depth encoding; (2) w/o \PVL{} disables Projection-Consistent Vision Lifting in the point branch; (3) w/o \CQP{} removes cross-modal query pruning and retains all 2D--3D matching during early refinement.

\subsubsection{Effects of the \RMDE.}
As shown in Table~\ref{tab:ablation_modules}, with \RMDE{}, the performance on the KITTI dataset is improved on all metrics. Specifically, RTE decreases from 0.12 m to 0.11 m, RRE decreases from 0.58$^\circ$ to 0.55$^\circ$, and Acc increases from 99.61\% to 99.70\%. This demonstrates that incorporating metric depth and ray-direction cues consistently improves image representations before cross-modal interaction.

\subsubsection{Effects of the \PVL.}
Without \PVL{}, RRE increases from 0.55$^\circ$ to 0.74$^\circ$, which is the largest degradation among all variants, while other metrics remain relatively stable. This suggests that Projection-Consistent Vision Lifting mainly benefits rotation estimation by improving cross-modal geometric alignment.

\subsubsection{Effects of the \CQP.}
Without \CQP{}, RTE increases from 0.11 m to 0.18 m and Acc drops from 99.70\% to 98.82\%, showing the most significant performance degradation. This indicates that suppressing unreliable matching during early refinement is crucial for stable correspondence learning.

\begin{table}[!htbp]
\caption{The effect of each design in our framework.}
\label{tab:ablation_modules}
\centering
\renewcommand{\arraystretch}{1.03}
\begin{tabular}{l|c|c|c}
\hline
Methods & RTE (m)$\downarrow$ & RRE ($^\circ$)$\downarrow$ & Acc. (\%)$\uparrow$ \\
\hline
w/o \RMDE{} & 0.12$\pm$0.11 & 0.58$\pm$0.72 & 99.61 \\
w/o \PVL{} & 0.12$\pm$0.14 & 0.74$\pm$0.87 & 99.46 \\
w/o \CQP{} & 0.18$\pm$0.22 & 0.65$\pm$0.51 & 98.82 \\
\hline
DPA-I2P (Full) & \textbf{0.11$\pm$0.12} & \textbf{0.55$\pm$0.67} & \textbf{99.70} \\
\hline
\end{tabular}
\end{table}

\subsubsection{Component Analysis of RMDE.}
Table~\ref{tab:ablation_rmde_cues} analyzes different ways of exploiting the same frozen UniDepthV2 prior in the image branch, while keeping the remaining framework unchanged. Directly concatenating the raw metric depth map provides only a marginal improvement over the no-depth baseline, reducing RRE from $0.58^\circ$ to $0.57^\circ$ and increasing Acc. from $99.61\%$ to $99.63\%$. This indicates that the metric depth prior itself is useful, but naive depth concatenation is insufficient to fully exploit its geometric information.

Introducing camera-ray direction into the depth representation further improves the performance, yielding an RRE of $0.57^\circ$ and an Acc. of $99.64\%$. This suggests that metric depth becomes more informative when explicitly coupled with camera geometry rather than treated as an independent scalar cue. Encoding the depth as a camera-aware surface point further reduces RRE to $0.56^\circ$ and improves Acc. to $99.66\%$, indicating that explicitly representing the 3D location associated with each image token provides more effective geometric supervision.

We then extend the single surface estimate to a local neighborhood by uniformly sampling multiple points along each viewing ray. The resulting variant further improves the performance to an RRE of $0.56^\circ$ and an Acc. of $99.67\%$, supporting the use of a local ray distribution instead of relying on a single predicted surface point. Removing confidence-aware aggregation slightly degrades the performance, with Acc. decreasing to $99.66\%$ and RRE increasing to $0.56^\circ$. This indicates that confidence provides a complementary benefit by reducing the influence of less reliable monocular depth predictions.

Finally, the full \RMDE{} achieves the best performance, with an RTE/RRE/Acc. of $0.11\pm0.12$~m/$0.55\pm0.67^\circ$/$99.70\%$. Overall, the results show that the gain of \RMDE{} does not come merely from appending an external depth prior, but from progressively structuring the prior through camera-ray geometry, local ray sampling, and confidence-aware aggregation before cross-modal correspondence refinement.
    
\begin{table}[!htbp]
\caption{Ablation study of RMDE components on KITTI. All variants use the same frozen UniDepthV2 prior when depth is involved, and the remaining framework is kept unchanged.}
\label{tab:ablation_rmde_cues}
\centering
\renewcommand{\arraystretch}{1.03}
\begin{tabular}{l|c|c|c}
\hline
RMDE variant & RTE (m)$\downarrow$ & RRE ($^\circ$)$\downarrow$ & Acc. (\%)$\uparrow$ \\
\hline
No depth prior
& 0.12$\pm$0.11
& 0.58$\pm$0.72
& 99.61 \\

Raw depth concatenation
& 0.12$\pm$0.12
& 0.57$\pm$0.70
& 99.63 \\

Depth + ray direction
& 0.12$\pm$0.12
& 0.57$\pm$0.69
& 99.64 \\

Surface-point encoding
& 0.12$\pm$0.12
& 0.56$\pm$0.69
& 99.66 \\

Uniform ray sampling w/o confidence
& 0.11$\pm$0.12
& 0.56$\pm$0.68
& 99.67 \\

Full \RMDE{}
& \textbf{0.11$\pm$0.12}
& \textbf{0.55$\pm$0.67}
& \textbf{99.70} \\
\hline
\end{tabular}
\end{table}

\subsubsection{Effects of the Pruning Schedule.}
Table~\ref{tab:ablation_pruning} shows that applying query pruning only in the first two refinement stages achieves the best performance, reducing RTE from 0.18 m to 0.11 m and RRE from 0.65$^\circ$ to 0.55$^\circ$ compared with no pruning. In contrast, applying pruning throughout all stages leads to inferior results, suggesting that later refinement requires more flexible local correspondence exploration rather than overly strong projective constraints.
\begin{table}[!htbp]
\caption{Ablation studies on query pruning strategies on the KITTI dataset.}
\label{tab:ablation_pruning}
\centering
\renewcommand{\arraystretch}{1.03}
\begin{tabular}{l|c|c|c}
\hline
Query strategy & RTE (m)$\downarrow$ & RRE ($^\circ$)$\downarrow$ & Acc. (\%)$\uparrow$ \\
\hline
No query pruning & 0.18$\pm$0.22 & 0.65$\pm$0.51 & 98.82 \\
Early-only & \textbf{0.11$\pm$0.12} & \textbf{0.55$\pm$0.67} & \textbf{99.70} \\
All-stage & 0.14$\pm$0.43 & 0.66$\pm$0.87 & 98.90 \\
\hline
\end{tabular}
\end{table}

\subsection{Efficiency Analysis}

We evaluate the efficiency of DPA-I2P on the KITTI dataset. All methods are evaluated under the same input setting, where the image is resized to $160 \times 512$ and the point cloud is downsampled to 40,960 points. Neural inference, including feature extraction, cross-modal correspondence refinement, and differentiable pose solving, is performed on an NVIDIA GeForce RTX 4090 GPU. Following our implementation setting, we use a batch size of 4 and report the average network size, GPU memory consumption, and end-to-end inference time per image--point cloud pair. Since the metric depth and confidence maps are pre-computed offline using the frozen UniDepthV2 model, this preprocessing cost is excluded from the reported online inference time.

The results are summarized in Table~\ref{tab:efficiency}. Classification-based methods such as Grid Cls. achieve lower computational and memory overhead, as they do not perform dense cross-modal correspondence refinement. Recent implicit correspondence learning methods require additional computation to establish and refine cross-modal correspondences, resulting in moderately higher resource consumption. DPA-I2P has a network size of 179.93~MB, uses 11.15~GB of GPU memory, and takes 36.81~ms for end-to-end inference. Compared with ICLI2P, these values increase from 175.92~MB to 179.93~MB, from 10.74~GB to 11.15~GB, and from 35.12~ms to 36.81~ms, corresponding to increases of approximately 2.3\%, 3.8\%, and 4.8\%, respectively. Meanwhile, DPA-I2P achieves substantially improved registration accuracy, as reported in Table~\ref{tab:main_results}. The relatively small increase in model size, GPU memory usage, and inference time indicates that the proposed RMDE, PVL, and CQP modules introduce only moderate additional overhead, while the overall inference efficiency remains within the range of recent implicit correspondence learning methods.

\begin{table}[!t]
\caption{Efficiency comparison on the KITTI dataset. Network size, GPU memory, and inference time are measured on an NVIDIA GeForce RTX 4090 GPU. The reported inference time includes feature extraction, correspondence refinement, and differentiable pose solving. Offline metric depth pre-computation is not included.}
\label{tab:efficiency}
\centering
\renewcommand{\arraystretch}{1.03}
\setlength{\tabcolsep}{4.2pt}
\begin{tabular}{l|c|c|c}
\hline
Methods 
& Network size (MB) 
& Mem. (GB) 
& Inference (ms) \\
\hline
Grid Cls.~\cite{li2021deepi2p}
& 100.75 & 2.39 & 11.20 \\
DeepI2P~\cite{li2021deepi2p}
& 100.12 & 2.01 & 7.55 \\
CorrI2P~\cite{ren2022corri2p}
& 141.07 & 2.88 & 13.75 \\
VP2P-Match~\cite{zhou2023differentiable}
& 146.42 & 17.17 & 30.63 \\
ICLI2P~\cite{li2025implicit}
& 175.92 & 10.74 & 35.12 \\
\hline
DPA-I2P (Ours)
& 179.93 & 11.15 & 36.81 \\
\hline
\end{tabular}
\end{table}
\section{Conclusion}
We propose DPA-I2P, a depth-aware implicit correspondence learning framework for Image-to-Point Cloud Registration. RMDE and PVL improve cross-modal feature representation using metric depth and projection-consistent vision cues, while CQP enhances matching stability by suppressing unreliable queries. Experiments on KITTI and nuScenes demonstrate superior performance, robustness, and cross-dataset generalization.

\section*{Acknowledgment}

\bibliographystyle{IEEEtran}
\bibliography{references}

\end{document}